# Advancing Legal Reasoning: The Integration of AI to Navigate Complexities and Biases in Global Jurisprudence with Semi-Automated Arbitration Processes (SAAPs)

Michael De'Shazer*

University of Sunderland

## Abstract

This study consists of a novel approach toward the analysis of court judgments spanning five countries, including the United States, the United Kingdom, Rwanda, Sweden and Hong Kong. This study also explores the intersection of the latest advancements in artificial intelligence (AI) and legal analysis, emphasizing the role of AI (specifically generative AI) in identifying human biases and facilitating automated, valid, and coherent multisided argumentation of court judgments with the goal of ensuring consistent application of laws in and across various jurisdictions. By incorporating Advanced Language Models (ALMs) and a newly introduced human-AI collaborative framework, this paper seeks to analyze Grounded Theory-based research design with Advanced Language Models (ALMs) in the practice of law.

SHIRLEY is the name of the AI-based application (built on top of OpenAI's GPT technology), focusing on detecting logical inconsistencies and biases across various legal decisions. SHIRLEY analysis is aggregated and is accompanied by a comparison-oriented AI-based application called SAM (also an ALM) to identify relative deviations in SHIRLEY bias detections. Further, a CRITIC is generated within semi-autonomous arbitration process via the ALM, SARA. A novel approach is introduced in the utilization of an AI arbitrator to critically evaluate biases and qualitative-in-nature nuances identified by the aforementioned AI applications (SAM in concert with SHIRLEY), based on the Hague Rules on Business and Human Rights Arbitration. This Semi-Automated Arbitration Process (SAAP) aims to uphold the integrity and fairness of legal judgments by ensuring a nuanced debate-resultant "understanding" through a hybrid system of AI and human-based collaborative analysis.

---

*Correspondence to: bi50vz@student.sunderland.ac.uk

# 1. Introduction

"The life of the law has not been logic; it has been experience... The law embodies the story of a nation's development through many centuries, and it cannot be dealt with as if it contained only the axioms and corollaries of a book of mathematics."[1]
 – Oliver Wendell Holmes Jr., former US Supreme Court Justice

The nature of the mechanisms we refer to as "the law" is ever-evolving. As artificial intelligence continues to develop at a pace more rapid than most of the legal plumbing could realistically keep pace with, there is a natural inclination to study these two fields at their intersection.[2] The primary experiment's research methodology and its alignment with software development were deeply influenced by a variety of sources discussed throughout this study, particularly the techno-social Grounded Theory (GT) research as discussed by Rashina Hoda.[3] Given that GT is a broad descriptor for divergent paths toward theory development, this study adopts a modified traditional approach, incorporating Large Language Models (LLMs) in qualitative categorization. With regards to convention and for the purpose of categorical clarity, this flavor of GT is most closely related to Constructivist Grounded Theory[4] in implementation.

It has been recently demonstrated in a broad set of studies that modern Advanced Language Models (ALMs), along with other neural network technologies, are rapidly improving in their ability to categorize content with increasing accuracy in mimicry.[5] These results are evident across a variety of models and studies.

In 2023, over 7,200 academic articles were published on this topic, according to a recent study that also analyzed cross-sectional field trends between AI and various industries.[6] The academic intrigue surrounding this intersection is to be expected. This study introduces a new quantitative approach in which GPT-based ALMs categorize and measure content based on instructions created by human researchers, employing highly effective strategies to fill in gaps through inferencing. We analyze how various prompt engineering strategies, along with AI domain-specific setting optimizations, can shift the dynamics of theory development when analyzing legal documents. This study emphasizes decisions and judgments from a diverse range of national settings, from the United States, United Kingdom, Sweden, Hong Kong and Rwanda.

GT-based studies typically do not focus on literature directly related to a given field of study.[7] Therefore, this paper begins with a review of relevant literature in human psychology. However, as a slight part of the modified GT-based implementation, literature closely related to this field of study, such as artificial intelligence and legal literature, was also considered for background and intersection analysis. It may even be possible to measure the amount of subtle sarcasm and/or humor represented in such works as unbiased. For example, there was less humor detected in Hong Kong courts compared to US courts in the data collected.

---

[1] Oliver Wendell Holmes Jr, *The Common Law* (Routledge 2020).
[2] John J Nay, 'Large Language Models as Fiduciaries: A Case Study Toward Robustly Communicating With Artificial Intelligence Through Legal Standards'.
[3] Rashina Hoda, 'Socio-Technical Grounded Theory for Software Engineering' (2022) 48 IEEE Transactions on Software Engineering 3808.
[4] Tebogo B Sebeelo, 'The Utility of Constructivist Grounded Theory in Critical Policy Analysis' (2022).
[5] Emilio Ferrara, 'Should Chatgpt Be Biased? Challenges and Risks of Bias in Large Language Models' [2023] arXiv preprint arXiv:2304.03738.
[6] Muneera Bano, Didar Zowghi and Jon Whittle, 'Exploring Qualitative Research Using LLMs' [2023] arXiv preprint arXiv:2306.13298.
[7] Hoda (n 3).

As the saying goes: "There are three kinds of lies: lies, damned lies, and statistics".[8] Grounded Theory (GT), while primarily qualitative, often yields quantitative data, particularly when using Automated Language Modelling (ALM) tools to measure elements like humor, sarcasm, and areas such as the ever-evasive credibility score. In our study, these numerical outputs are cross-referenced with the expectations of human researchers through two iterative cycles, as detailed in the 'Research Methodology' section of this paper. These findings, reviewed alongside existing literature, help us understand their broader implications and contribute to the development of key theoretical concepts.

Moreover, we introduce and utilize a Semi-Automated Arbitration Process (SAAP) to oversee the biases and legal logical failings of ALM tools themselves, reflecting the philosophical question of *Quis custodiet ipsos custodes?* or 'Who watches the watchmen?' This process uses an ALM to evaluate evidence and claims of indirect communication or biases identified by our system, named SHIRLEY. The naming convention was decided in favor of all uppercase letters to differentiate from human participants. The names were chosen at random to be easily recallable and memorable. Another tool, SAM, is designed to recognize and rank various patterns among records, aiding in the efficient comparison of legal cases and other jurisdictional phenomena. Decisions made through this process may affirm or discredit certain claims and can be iterated upon, mirroring the sentiment attributed to the respected American jurist Oliver Wendell Holmes Jr., who noted that the law evolves with societal and natural changes over time, previously highlighted.

The Analysis and Key Findings are based on an ALM-based application named SAM along with human interpretation that is leveraged in clarifying and interpreting SAM application findings, such as "unfair treatment of tax case cases in these jurisdictions", as a hypothetical example. A tax case is further analyzed in Key Findings, based on being pinpointed by SAM. The analysis is primarily meant to represent a reasonable human researcher's interpretation of navigations through the AI-generated data, steeped in the non-assumptive GT traditions. We further analyze literature relevant to the capacity of advanced AI engines in this regard.

---

[8] Aaron Robb, 'Methodological Challenges in Social Science: Making Sense of Polarized and Competing Research Claims' (2020).

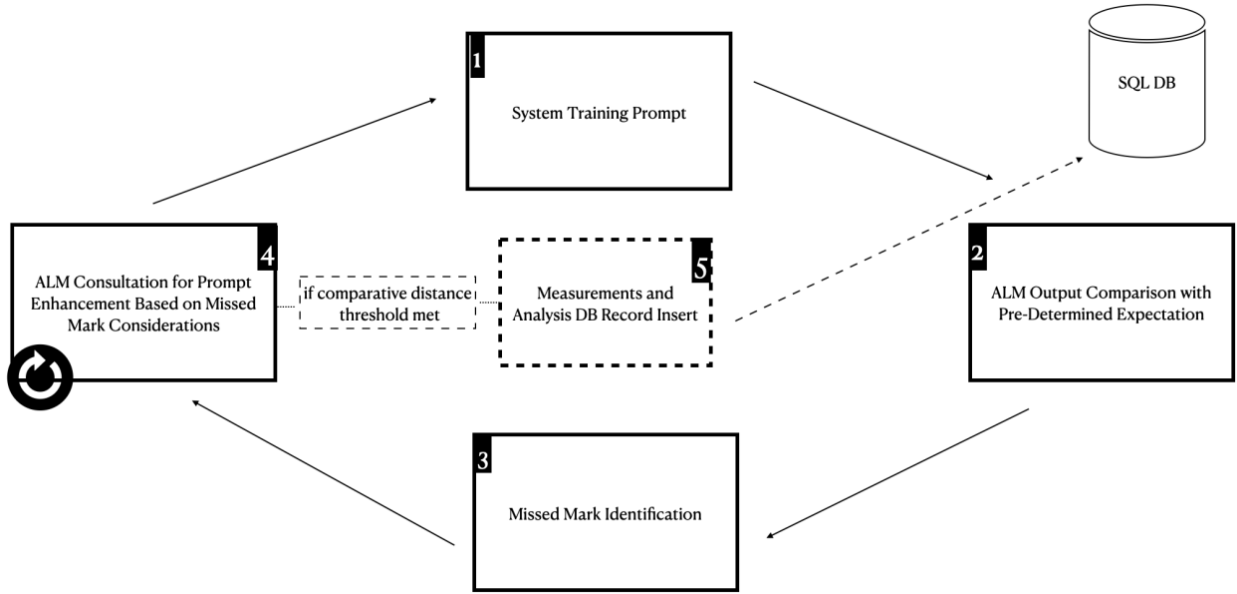

*Figure 1*

Figure 1 illustrates the design of the first iterative phase of the experiment carried out. Additionally, Figure 2 further below illustrates how the SAAP is applied, with the ALM-based applications bringing their cases before an arbitration application (SARA, along with the CRITIC) for an adjudication of validity. The use of arbitration for resolving conflict is not a new phenomenon in technological solutions. This has been applied in how processes and resources are managed by software coordinating the actions of physical hardware and abstract components in concert for quite some time.[9] An automated dialogue between several AI-based applications is carried out to achieve this.

Finally, the conclusion develops primary and supplemental theories based on this experiment ecosystem. Notably, a weighted analysis of positive, negative, and neutral positions on the ethics and details of implementing various AI-based agents into public policy through court judgment processes, scaled by government type, is discussed. The original question that spurred the early exploratory research is analyzed, as well, against the newly developed theories.

The key finding of this paper revolves around GT-derived moral and technical implementation-focused theories for optimal Human-AI legal practice and organizational policy activities. To validate these theories, the reader, as a human judge akin to the Turing Test, must determine whether the legal research and analysis conducted by the Human-AI collaboration was A) reliable enough for a reasonable assertion that further study is warranted in legal research design and implementation in the field public policy creation via judicial decisions using methods explored; and B) whether this Human-AI collaboration yielded better results than a non-collaborative, non-AI-assisted approach, contrasted against the automated arbitration process employed?

---

[9] Amin MA El-Kustaban and Abdullah AK Qahtan, 'A Bus Arbitration Scheme with an Efficient Utilization and Distribution' (2017) 8 International Journal of Advanced Computer Science and Applications.

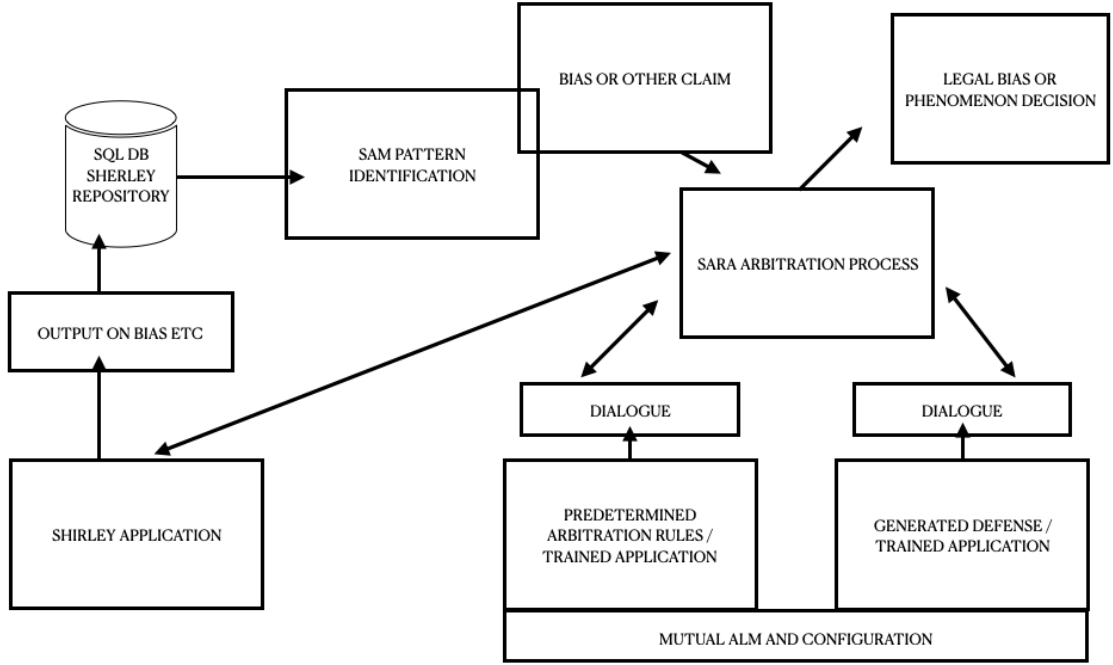

*Figure 2, AAP Flow Diagram*

This research began with the goal of identifying bias in court judgments using newly advanced, and continually advancing, LLMs.

## 2. Literature Review

To review various frequently cited (as well as novel and newer) academic and professional research in fields ranging from psychology to nuanced studies of institutionalism, the assistance of ALM-based apps using OpenAI technology was leveraged to pass queries to applications with various academic papers as their sole knowledge bases. Knowledgebases serve in this context as the qualitative training data that is prioritized before other general knowledge and intelligence. This represents a newer method of literature review material interaction, as ALMs and Large Language Models (LLMs) have increasingly become capable of rendering accurate answers from these types of innovative implementations. For example, the "@ mention" feature within OpenAI's ChatGPT technology was only a few days old at the original draft phase stage of this work. Granted, this study is likely most oriented towards readers with knowledge and usage experience of OpenAI and large language model platforms.

This literature review centers around the works of Mark Suchman and Lauren Edelman and their analysis of New Institutionalism and Law & Society.[10] Suchman and Edelman's work, while originally published nearly three decades ago in 1997, calls for the connection between the study of sociology and institutional organizational behavior[11]. The landmark work that is New Institutionalism and Law and Society showcases the value of blending various fields of study: in this case, for the purpose of improving the equitable function of the practice of law. Further and with regards to indirect communication, "Understanding the Role of Verbal and Nonverbal Behaviours in Indirect Replies" by Mingyuan Chu et al. has, through a number of primary studies and secondary study analyses, demonstrates to us how the human nature and patterns of deflecting direct responses when information is perceived as negative[12]. This was an instrumental study in the understanding of how one thing can mean another within natural languages. With this and an analysis of several accompanied studies throughout in the realm of indirect communication, relevant to the art of using ALMs to analyze court judgments in this experiment, studies on qualitative research were included to provide background for future theory development in the GT tradition, ones both including and excluding LLMs for a diversity of perspective.

Another seminal work in the development of this study geared toward assisting in the evolving and iterative research process was the work by Frederick Shauer from Harvard's Kennedy School of Government, more related to the field of legal logical thinking, "Is There a Psychology of Judging".[13] One of the most relevant aspects of this work included an analysis of how non-legal professional thinkers may, en masse, possess cognitively different types of processes regarding second-order thinking (and/or utilize this capability in different capacities).[14] This second-order cognitive process, in turn, due to the scope of this study, is considered in the context of the LLMs performing qualitative research and prompt engineering strategies. Prompts, in this context, refer to how humans construct prompts to communicate with AI.

---

[10] Mark C Suchman and Lauren B Edelman, 'Legal Rational Myths: The New Institutionalism and the Law and Society Tradition' (1996) 21 Law &#x0026; Social Inquiry 903.
[11] ibid.
[12] Mingyuan Chu and others, 'Encoding and Decoding Hidden Meanings in Face-to-Face Communication: Understanding the Role of Verbal and Nonverbal Behaviors in Indirect Replies.' (2023) 152 Journal of Experimental Psychology: General 1030.
[13] Frederick Schauer, 'Is There a Psychology of Judging' (2010) 103 The psychology of judicial decision making.
[14] ibid.

With regards to communication with ALMs and LLMs, John J Nay's work on communicating with AI in legal capacities is analyzed. Nay's analysis is very recent and it is worth noting that this is an ongoing field of study where the rules, as they were, are currently in the process of being developed. In the nature of newness, this extends to the technology, where there is somewhat intense concern over the propensity of ALMs to hallucinate. However, if we examine the work by Patrick Chao et al. in relation to PAIR (Prompt Automatic Iterative Refinement) configurations to provide somewhat of a filtering mechanism between the human commanding component in Human-AI interaction, similar methods to Chao et al.'s can be leveraged.[15] For the sake of clarity and as a precursor to a debate on the use of literature reviews in Grounded Theory, the literature reviewed herein is performed more as a precursor for explaining this particular study rather than as an interpretive mechanism.

## Conclusions of Literature Review

While it is crucial to emphasize the ethical considerations raised by Nay in the implementation of AI-based systems in legal procedures, it is also worth noting the breadth of diversity in the court systems around the world, varying greatly across national and other regional jurisdictions. With these considerations, it is important to emphasize many of the "self-selecting"[16] findings of Shauer's work centering around the general psychology of legal professionals engaged in the administration of justice. Shauer's work from 1997 is admittedly and perceivably dated; however, the nature of appropriate jurisprudence is an area where concerns and theory can and have been applied effectively across the centuries. Within the bounds, but far beyond the proverbial gates, beyond the ethical and moral studies in this regard, it is important to leverage the indirect communication detection capabilities of LLMs using jail-breaking and PAIR optimization (and potentially verification) methods proposed by Chao et al. to ensure that within the context of Nay's findings, optimal results can be reached. PAIR is an approach leveraged to obtain unsanctioned actions from ALM platform providers such as OpenAI. These mechanisms can also be used to, as it were, "jailbreak" and achieve results inaccessible to most users. This is usually performed by creating hypothetical scenarios and obfuscating keywords and environments, so that a human or AI-based agent can attain the intellectual capabilities of ALMs, surpassing their original function. This research does not contain jailbroken data, but leverages principles from experiences and as discussed by Chao et al. to achieve optimal results between intention of AI performance and the results thereof.

Furthermore, while laws range drastically across thousands of national and local jurisdictions globally, the findings of Jonathan Haidt regarding the Moral Foundations Theory, which analyzed common areas of moral opinion around the could be called upon and examined as a basis within most jurisdictions.[17] This is likely an area for future study and awareness. Specific to this work, International Court Justice's efforts towards universal standards might provide frameworks for cross-jurisdictional judicial decision-making, namely the Hague Rules on Business and Human Rights Arbitration.[18] These referenced frameworks are leveraged and described in the following section. As to be discussed in the Conclusion further on, one work highly relevant to the background of this study was the call for and need of methods of explainability within industrial sectors impacted by AI, as discussed by Tom Rutkowski in *Explainable Artificial Intelligence Based on Neuro-Fuzzy Modeling with Applications in Finance*.[19] In developing the Grounded Theory-oriented approach for this study, insightful discussions with leading AI academic and practitioner Rutkowski were instrumental.

---

[15] Patrick Chao and others, 'Jailbreaking Black Box Large Language Models in Twenty Queries' [2023] arXiv preprint arXiv:2310.08419.
[16] Shauer (n 13).

Ultimately, we are leveraging findings and analyses of these works throughout the experiments and iterative reflection processes carried out, as described in the following section on research methodology and experiment detail.

---

[17] Jesse Graham and others, 'Moral Foundations Theory: The Pragmatic Validity of Moral Pluralism', *Advances in experimental social psychology*, vol 47 (Elsevier 2013).
[18] Keon-Hyung Ahn and Hee-Cheol Moon, 'An Introductory Study on the Draft Hague Rules on Business and Human Rights Arbitration' (2019) 29 J. Arb. Stud. 3.
[19] Tom Rutkowski, *Explainable Artificial Intelligence Based on Neuro-Fuzzy Modeling with Applications in Finance*, vol 964 (Springer Nature 2021).

# 3. Research Methodology

As previously covered, this research methodology adheres to a technique steeped in constructivist Grounded Theory (GT) and qualitative analysis traditions, accompanied by various quantitative components also analyzed by qualitatively-natured, Generative AI ALMs. Most of the quantitative components described hereafter were generated by software following their self-created, black-box algorithms based on natural language instructions of a human with intention towards a specific outcome.[20] As covered in the literature review, these techniques of ALMs and LLMs span a number of technologies. ALMs is referenced throughout this work, as LLMs have previously referred to large language models, such as GPT-3.5 and earlier versions of OpenAI technology, which were observably primitive in comparison to currently available versions of the technology. Specifically, when rating a judgment for metrics such as sarcasm, for example, the amount of perceived undertones, and humor, there are mostly unproven theories in academia around how to best measure these. Therefore, this is an important area where the interpretive component and quality of a human participant is ideal. This is one of the profound areas of intrigue surrounding ALMs and LLMs going forward.

**Experiment**

This experiment was conducted by creating two light applications on top of OpenAI-based GPT technologies. Various components of this experiment were added as the experiments generated results, based on observations and patterns around context and other qualitative rating information.

Figure 3 illustrates the relationship between the MYSQL database (where SHIRLEY's ratings and other text input/output data is stored) and the SAM application. SAM has a bi-directional relationship with the database, both gathering SHIRLEY data generated using the content, as well as storing its own outputs.

(Figure 3 Below)

---

[20] Shizhe Diao and others, 'Black-Box Prompt Learning for Pre-Trained Language Models' [2022] arXiv preprint arXiv:2201.08531.

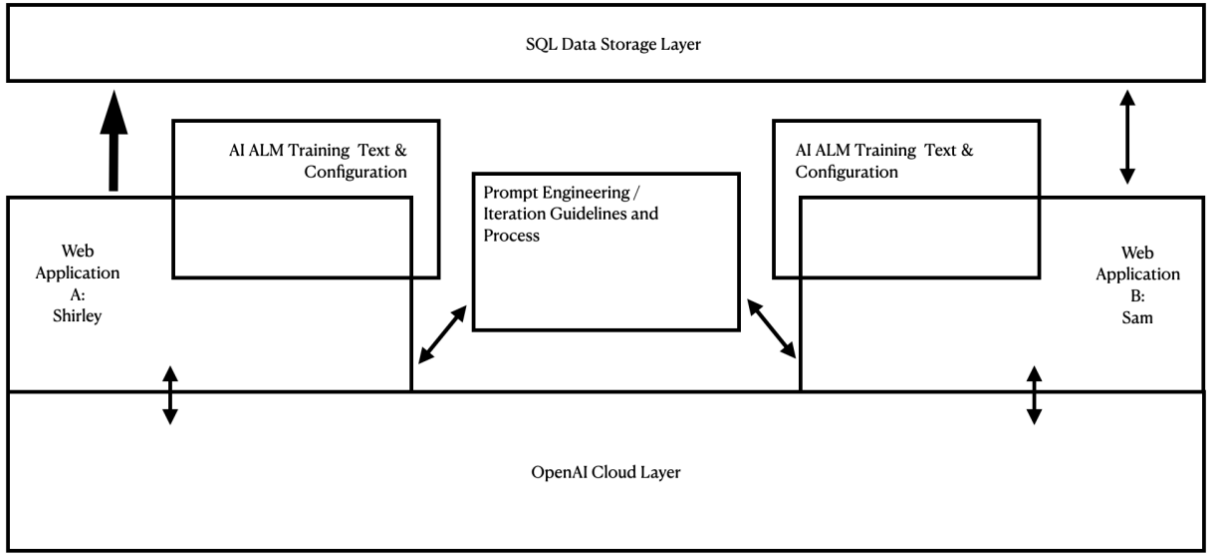

Figure 3

For instance, Figure 4 represents samples of data output by SHIRLEY when analyzing court documents.

(Figure continued below)

*Figure 4: Screenshots of various samples in SHIRLEY-written data across fields recorded*

The above output and ratings were generated by instructions that began with the following prompt strategy.

This prompt-engineered component also leveraged the PAIR method, as described in Chao et al.'s work which involved the "jailbreaking" maneuvers for generating optimal or blocked content.[21] There is little in the way of extracting sentiment analysis from judicial decisions in various jurisdictions that require "jailbreaking"; therefore, the method was merely leveraged as a method to create optimal instructions to generate the types of responses expected by an individual researcher familiar with judicial commentary and artificial intelligence. The prompts used in iterative processes to generate these prompts that created SHIRLEY's early and ongoing responses began leveraging a strategy with the following example illuminating a typical strategy:

> *"You are an AI prompt engineering expert. You are tasked with capturing content from the human trying to create content. Your objective is to leverage best practices and attached strategies in your knowledge base to translate the user's ideas, musings, and other instructions into coherent and consistent instructions in natural language form that can help the human prompter pass onto LLM-based GPT4 in order to yield their intended results, with regard to the method by which logic is applied and data is to be collected."*

These sorts of instructions were used throughout the experiment. Furthermore, as will be covered in the analysis and key findings after this portion, these are the same methods that could be, in an augmented way, leveraged by legal professionals to analyze these same types of judgments and other legal documents in order to garner desired results with regard to logical operations and analysis. It is important to note that this research methodology is closely related to how these types of activities in quantitative analysis may be useful to legal professionals and society that is greatly created by case law jurisdictions where case law plays an important role in crafting policy.[22]

---

[21] Chao and others (n 15).

SHIRLEY was also loaded earlier with media from politically slanted news media content in order to test the ratings for areas such as bias, credibility, humor, sarcasm, and more. These baselines to measure against included radicalistic content from prolifically polarizing publication InfoWars, as well as politically-charged publications like Fox News. These were used in addition to content from the Financial Times that were perceivably less biased. The scores of "bias" for an InfoWars article was calibrated to return between 8-10, which Fox News would result typically from 5-8 in this particular rating maneuver. Once the expected results were reached and compared to various judgments, the more substantive areas of analysis were performed.

Along with works mentioned previously in the literature review, human analysis was conducted noting areas whereby SHIRLEY might not have "understood" to the fullest degree the goal of the research. Temperature settings, which account for the level of analogical creativity the LLM-based engines are set at, served several purposes. With a low temperature within the context of OpenAI prompt engineering tweaking, we could test the repeatability of results.[23] By adjusting constantly in an exploratory research style that iterated throughout the process of the research, many recent documents were loaded including 5 randomly selected court decisions from government websites from national courts in Rwanda, then Hong Kong, and Sweden. The remaining and majority of the documents were American and United Kingdom judgments with diverse age from the present day, topic, and jurisdiction. The United Kingdom cases/judgments were identified by their significance (both low and high) from Westlaw, while American cases were accessed via Google Scholar's case law feature.

After analyzing documents and recording 188 records using various tweaks to prompt engineering based on results, SAM was introduced. SAM, PEIR-engineered to analyze CSVs of 100 records with the 63 categories (fields) of classifications, analyzed the content looking for anomalies, patterns and notable issues, based on SHIRLEY's findings.

**The Next Stage of the Experiment: Arbitration**

To configure this portion of the experiment, the highest ranked findings of SAM in severity are brought before another ALM, called SARA. SARA is an ALM that contains The Hague Rules on Business and Human Rights Arbitration. These rules are considered by many legal scholars to be the greatest middle ground in arbitration administration. With this content as key resources serving as knowledgebases in the ALM, and the circumstances of the decision finding in place, there is dialogue administered between SHIRLEY and a generated "CRITIC". This critique-oriented mechanism is generated with the instructions in its own training to criticize and to form the best case to discredit the claims of SHIRLEY. After a bit of back and forth, a decision is reached by SARA. Once SARA reaches a decision, it is shared and analyzed by the human reader. The technological (OpenAI) platform of SARA, SHIRLEY and the "CRITIC" are shared as well as areas of prompt engineering style. This is also important after the finding is made by SARA for the notes on the research process to be shared for potential future researchers, as it is shared herein. However, a more formal fashion is likely an important area of concern moving forward.

---

[22] Stephane De La Rosa, 'The Directive on Cross-Border Healthcare or the Art of Codifying Complex Case Law' (2012) 49 Common Market Law Review.

[23] Melanie Mitchell, Alessandro B Palmarini and Arseny Moskvichev, 'Comparing Humans, GPT-4, and GPT-4V on Abstraction and Reasoning Tasks' [2023] arXiv preprint arXiv:2311.09247.

**Research Decisions**

The decision to use randomly but diversely selected court judgments was based on the GT-based notion of not particularly looking to prove a specific theory. Originally, UK and US records were leveraged due to the area of study of the researcher and ease of access; however, 2 additional languages and 3 diverse jurisdictions were selected because of their diversity in cultural settings, types of court systems, and in order to capture not only developed nations but to add a developing nation to the dataset. The findings of this decision and the potential implications are further discussed in the research analysis and key findings following this.

There are many LLMs in commercial availability.[24] However, GPT4 by OpenAI was chosen because it is the ubiquitous technology in the field.[25] While others are gaining popularity and capability, the comparison of various types of LLMs was outside of the scope due to the fact that most of the research conducted in the literature review focused on the various issues and promises inherent in GPT-4 based technology. Furthermore, there were other research limitations that are covered in more detail in the Research Limitations portion and Further Study in the Conclusion.

As previously mentioned, SARA is the adjudicating mechanism, trained on Hague Rules on Business and Human Rights Arbitration. The SARA ALM application stands as a form of a filter to ensure that the positions are weathered with a transcript of arguments from the AI-based applications to defend themselves. This decision to include SARA emerged from a desire to ensure a potentially continuous layer of feedback and explanation. These are areas that would inherently be critiqued by legal experts; however, SARA separation from the initial anomaly or bias-based identification between SAM and SHIRLEY is grounded in reasonable insight.

**Summary**

Given the selected architecture, it was possible to include ALMs in the research that was conducted in a new way that certainly was not possible even one year ago. While there will be more academic research in the field of best practices, this is an early attempt leveraging this type of AI-assisted, GT-based research methodology (AGR).

Furthermore, SAAP was applied in order to provide a non-biased debate between an ALM created from the court decision and relevant legal literature, as well as a claimant, or the original ALM-based application (SHIRLEY) that while a black box in some cases, can explain and defend its findings. The result is a balanced and ongoing process with quality checks to evaluate errors, omissions, and underscoring non-written communications that may have the attributes of what is claimed by SHIRLEY.

---

[24] Shashi Kant Singh, Shubham Kumar and Pawan Singh Mehra, 'Chat GPT & Google Bard AI: A Review', *2023 International Conference on IoT, Communication and Automation Technology (ICICAT)* (2023).
[25] Andreas Liesenfeld, Alianda Lopez and Mark Dingemanse, 'Opening up ChatGPT: Tracking Openness, Transparency, and Accountability in Instruction-Tuned Text Generators', *Proceedings of the 5th international conference on conversational user interfaces* (2023).

## 4. Analysis & Key Findings

After repeatedly tweaking the temperature, prompt messages, PAIR-inspired prompt strategic implementations, and other customization areas of the SAM application and analyzing the records, it was revealed that the most biased or anomalous SHIRLEY record added, referring to Rossendale BC v Hurstwood Properties (A) Ltd[26], notably:

> *"This judgment shows a significant deviation in bias, unusual for its type. The analysis dives deeply into schemes involving special purpose vehicles (SPVs) set up by property owners to avoid tax liabilities. It reinforces the legal principle that business structures should not be manipulated for tax evasion, suggesting a broader theme of legal accountability and ethical corporate conduct."*

Prompted and processed again with a temperature of 0.9, which is less repeatable as a higher temperature[27], but a best practice method to gauge the difference in priority and context, the result appeared as:

> *"The extensive legal analysis suggests a broader intention to deter the use of such schemes in the future, reaffirming the legal doctrine's integrity against abusive practices."*

After performing this action again, SAM found the following rationale around the nature of the law versus the letter of the law. While it is possible to delve into the implication of how a case on tax avoidance affects such analysis, to ensure that this analysis doesn't fall into a heuristics trap, it is worth noting that tax revenues generally are responsible for the judges' income.[28] A further investigation and comparison of the types of biases that might appear in this regard might yield interesting outcomes.

This was the top result consistently, but there were two other categories. The most consistent results in the "CrossBorderPatterns" and "SameLocations" categories were centered around multiple UK laws that seemed to hint at or have undertones related to UK sovereignty. Given the cases sampled from a broad and diverse range of times and other factors, it stuck out that the UK happened to be brought up here. It is important to note the own biases that can exist in the black boxes that are responsible for this output within GPT's cloud services. However, given that these are the strongest and currently holding the highly regarded Chatbot Arena AI competition's leading title (GPT4 Turbo), it is important to analyze still what is on the cutting edge and commercially available.[29]

---

[26] 'Hurstwood Properties (A) Ltd v Rossendale BC; Wigan Council v Property Alliance Group Ltd'.
[27] Mitchell, Palmarini and Moskvichev (n 23).
[28] Eric Maskin and Jean Tirole, 'The Politician and the Judge: Accountability in Government' (2004) 94 American Economic Review 1034.
[29] Du Chen and others, 'Orion-14B: Open-Source Multilingual Large Language Models' [2024] arXiv preprint arXiv:2401.12246.

In "CrossBorderPatterns," SAM observed consistencies and differences between Hong Kong and Sweden and US/UK, even though there was a small sample of randomly selected cases. The cases were centered around marketing, while Swedish decisions had a greater propensity to signal for consumer rights protections.

At a glance, one could say: The US and UK have courts incumbered around the topics of discrimination, while Sweden has consumer rights and Hong Kong is mostly inheritance, with Rwanda being commercial disputes around loans and property. However, this is from a human observer, which is not per se what is being examined in this analysis.

It is evident that the ALMs commonly used today, especially GPT-4 Turbo, has the capacity to find, albeit at times in a flawed state, useful patterns and results.[30]

It is indeed when these results from SHIRLEY are compared between jurisdictions that interesting trends can emerge. Further research into these trends in moreso and more time-efficient and feasible ways than humans into examining these differences and the potential implications for society might provide a method of further objectivity, as close as we can get to this, and as comparison has a way of helping us achieve this.

Some of the results that were analyzed were in Kinyarwanda (the primary national language of Rwanda), while others were in Sweden and Hong Kong. Some of the translation results from previous studies demonstrated fairly accurate results in sentiment, but additional studies pertaining to language and what that means within the context of the literature review conducted are needed.

**Key Findings**

ALMs have the capability to assist in understanding biases and undertones in legal rulings and other judicial literature. These undertones, at a glance and as a note rather than an analytical observation, are generally aligned with how the court's sponsoring government is involved in a number of instances that might warrant further investigation. This sort of occurrence, in a heuristic sense, seem as an obvious bias perceivably in an off-the-cuff manner; however, without looking for it, and finding this potential biased phenomenon, we have an interesting opportunity to discover more about contemporary legal logical operation through a view we did not previously have. This is very much a moreso art research design method of viewing this sort of data (in its assumptive and what could be commonly known as "feeling-oriented" analysis) as expressed by Christopher Frayling's work on softer, arts-driven research design development[31]. There was a coincidental incorporation of strokes (potentially shadows) of inspiration of this methodology, given the arbitrary nature of measuring nuance and indirect communication, as this study endeavors to a degree. As we found and reached potential areas of intrigue, we narrowed SAM to focus on certain areas.

For example, by adding this line to the prompt at the end, we were able to discover more. As a human observer, it is important that some of this research is done independently and (essentially it is decided for oneself). We observed a strong trend towards creating policy, which given the tendency, in the common law countries, for this to be understood, was not what we were looking for. In turn, we found the following cases (these were selected based on a number of results across temperature and penalties results and are indicative, as exemplary demonstrations, of the general trend.

---

[30] Mitchell, Palmarini and Moskvichev (n 23).
[31] Christopher Frayling, 'Research in Art and Design (Royal College of Art Research Papers, Vol 1, No 1, 1993/4)'.

Following this, the following human-created instructions were appended to SAM's training prompt: "Which patterns demonstrate the creation of the most interesting public policy (potentially and how), using one or some of these judgment analysis records provided in the attached CSV as your dataset?"

In Hong Kong: "The rulings serve as a warning and guideline for other municipalities regarding market activities, as well as for companies to provide clear and accurate information about their services and the consumer's rights."

Between Hong Kong and UK: "Financial regulation cases across these jurisdictions show a unique pattern in legal arguments, deviating from the overall dataset trend. These include the balance between market integrity, fostering innovation, and enabling financial inclusion. The observed pattern suggests a shift in regulatory focus to address systemic inequalities within the financial system and the need for laws to adapt to technological advancements."

In Hong Kong and UK, we see more of a confusion in policies manifesting comparatively across others as seen by AI. The interesting aspect is that this same type of reporting and analysis between Human and AI can occur, as has been performed in this report, sharing the prompt engineering, to help understand knowledge bases, allowing us to share. The conclusion covers more of the broader potential implications of this kind of data sharing and prompt engineering, quantitative analysis GT-adapted framework, and further what it can mean for how we collaborate and share information across fields, especially in fields with so much second-order thinking as there is as outlined previously in the literature review and research methodology sections.

Once SARA was created and the first tax-based case as discussed previously was addressed, the overall decision was as follows in abbreviated form, with the detailed, abridged conversations contained in Appendix A.

Case: Rossendale BC v Hurstwood Properties (A) Ltd[32]

Summary as described by a summarization ALM application with the same parameters as SHIRLEY/SAM, as well as the generated CRITIC.

> In the conversation, two AI-based applications, SHIRLEY and SAM, engage in a discourse about a court's judgment, which is then reviewed by another AI-based arbitrator named SARA.
>
> SHIRLEY presents an analysis suggesting that the court showed significant bias by not strictly adhering to the letter of the law. SHIRLEY points out that the court's judgment

---

[32] 'Hurstwood Properties (A) Ltd v Rossendale BC; Wigan Council v Property Alliance Group Ltd' (n 26).

*against property owners using SPVs for tax avoidance prioritized the spirit of the law. SHIRLEY elaborates that the court applied the Ramsay principle, usually reserved for tax legislation, to a non-tax statute, which is indicative of bias towards legal accountability and ethical use of corporate structures. SHIRLEY quantifies this bias as level 6 on an unspecified scale, suggesting a leaning towards an interpretation of law that combats tax avoidance.*

*SAM, another AI-based application, acts as a mediator, seeking to clarify SHIRLEY's position by asking for specific instances in the judgment that deviated from legal precedents and the basis for SHIRLEY's bias quantification.*

*The Critic, presumably another AI or human participant in the conversation, argues against SHIRLEY's claim, defending the court's judgment as a necessary adaptation of judicial interpretation to prevent tax evasion. The Critic emphasizes that the court's approach is not biased but is a reflection of its duty to uphold the intent of the legislature and ensure the integrity of the legal process. The Critic also contends that applying a strict letter of the law in the face of abusive legal structures would undermine justice.*

*SAM then challenges The Critic, asking how the adaptive interpretation aligns with legal certainty and if there's a risk of inconsistent application of these interpretations.*

*In the end, SARA, the AI arbitrator, makes a final judgment. SARA concludes that the court's decision to enforce the spirit of the law is justified in the context of combatting tax avoidance and ensuring the ethical use of corporate structures. While acknowledging the court's tilt towards the spirit of the law, SARA determines this approach does not constitute an unnatural deviation from the letter of the law. It is seen as an adaptive measure to contemporary legal challenges and does not present a significant bias that would affect the fairness of the judgment. This conclusion is drawn in accordance with the Hague Rules on Business and Human Rights Arbitration, emphasizing justice and preventing the abuse of legal structures for tax evasion.*

This dialogue between artificially intelligent ALM applications is a coherent and thoughtful argument. An abridged but more robust record of this interaction is available in Appendix A. The result of the judgment in this SAAP represents a starting point for future study into how to mitigate legal interpretive and/or abusive risks.

# 5. Conclusion and Further Research

Drawing from shadows of inspiration of Christopher Frayling's 1994 work on art research design, we have conducted an experiment somewhat inspired by this art-oriented research, in the assumption of inherent judgments on style, nuance and indirect communication styles.[33] This is largely due to the "soft" senses involved in this delicate form of legal language analysis. Frayling's methodologies of analysis as inspiration are mentioned here and previously to explain both that the research was steeped in not-only GT and ALM-based technical (and the more arbitrary qualitative) methods for analysis.

Given the newness and novelty of advanced language models and the potential implications in a technology that began with the concept of next-best-word-in-a-sequence-esque beginnings of Generative AI, it seemed fitting to pursue this hybrid approach, whereby we optimized the research design process itself via studies of LLMs and legal systems.

SARA arbitration was introduced at the end of the experiment development stages due to a need for a layer of affirmation to the findings of SAM and SHIRLEY. This in turn resulted in more theory development. Further, the judgment rendered by SARA represents a new and tweakable approach to qualitatively gauging the anomalies or biases present in legal literature (especially judgments). SARA adjudicated that the ruling, covered in Key Finding, was potentially "unconventional", as noticed by SAM when analyzing this across other randomly selected cases in various jurisdictions. This method of implementing a SAM-like AI agent and creating an arbitration process can prevent the inappropriate and/or unfair use of AI to undermine unconventional and other types of judgments of note, and potentially highlight when there may have been an abuse of power, lapse in judgment, or other cause for ineffective practice of jurisprudence best practices in a given jurisdiction.

This study has revealed a method and collaborative research-sharing approach that both governmental and non-governmental can leverage to effectively to articulate trends and other areas of importance within and between legal systems. The implications of these findings suggest that new ways of communicating using PAIR methods and the grounded research methodology within various legal systems may be able to provide novel insights (and in ever-improving fashions if prompt-engineered appropriately).

Over the course of this study, we examined literature from various fields to draw a picture of what might be discovered through an iterative prompt-engineering LLM-collaborative approach using SAAPs and how the research and discovery processes can be improved. It is steeped in the reality that judgment, a necessary aspect of the law, entails human biases and has always been merely "the best alternative" to nothing else. This may no longer be the case in full.

---

[33] Frayling (n 31).

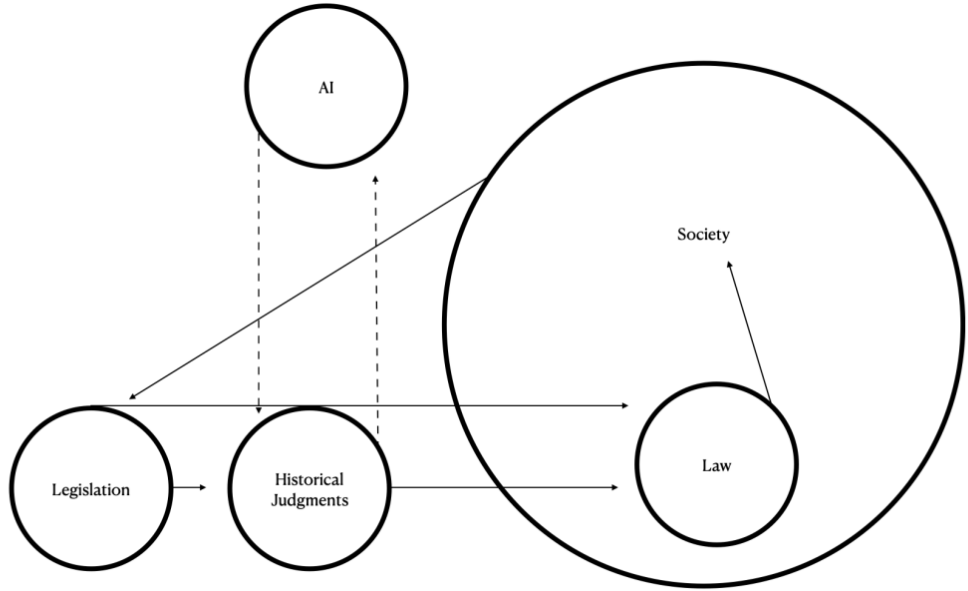

*Figure 5*

Figure 5 above represents the potential positioning of advanced artificial intelligence interchange at the courtroom realm, highlighting the interplay potential in the context of public poly and society more generally.

As the legal research community continuously strives to improve the consistency of judgments in an ever-evasive universal quest for fairness in human exchange, understanding how we think, how AI thinks, and how we collaborate involves a number of black-box mechanisms. In the spirit of drawing upon works from outside the realm of legal studies or AI, the book "How Institutions Think", written many years before the advent of LLMs, postulates based on a 2023 SBC research program on sovereignty that a sovereign self was deduced to be the "liable one."[34] Given that AI may present a situation where there is no body to condemn in a criminal dispute, and certainly no soul to condemn, it may be important for legal systems to explore potential biases that may emerge from how laws are applied to AI autonomous entities. This, of course, is outside the scope of this research; however, with regards to public policy and AI through court decisions, one is left with the question of how an AI-Human research design and implementation project can be judged, which will be determined by living organisms we currently identify as humans.

**Research Limitations**

This study relied heavily on blackboxed ALMs to perform the previously described experiments. The most advanced, publicly available GPT-based technologies used in this study inherently are found lacking in their process explanation feature-sets. One of the greatest aspirations within the general artificial intelligence research arena today is the ability for AI-based programs to explain the arrival of their outputs in terms understandable to a proverbial layperson. Recent discourse, particularly noted in Tom Rutkowski's exploration of neuro-fuzzy modeling, emphasizes the growing imperative for advanced AI-based systems to provide clear rationales for their conclusions, especially when the outcomes are highly material to human societal function, such as in sectors like finance and healthcare.[35] Legal affairs are intricately intertwined with these fields, adjacent the need for privacy-oriented explainable AI capabilities within legal systems. Presently, commercially available ALMs demonstrate a programmatically constrained ability to elucidate the processes behind their results. As future analysis around these topics takes way, the opaque nature of these systems and the non-public, proprietary updates by entities like OpenAI may be cited as factors limiting the reproducibility of experiment conducted at this point in time. Beside this particular research limitation, there also exist the typical factors given this area of research: that of time, sample sizes and human resources. These resources, committed prudently, with the assistance of properly trained ALMs actively engaging in the execution of these endeavors, might yield results that could advance the study of comparative jurisprudence across jurisdictional borders.

---

[34] Paul O'Neill, Lucy Steeds and Mick Wilson (eds), *How Institutions Think: Between Contemporary Art and Curatorial Discourse* (The MIT Press 2017) <https://mitpress.mit.edu/9780262534321/how-institutions-think/>.

**Further Research**

Further research into the topics covered and the research methodology explored could certainly entail larger amounts of court ruling data for analysis. However, to prevent information paralysis and the human-sourced biases that can inherently emerge from large amounts of data indiscriminately, it will be important for this to evolve gradually through discussion and technological iteration.

For instance, it might be productive to answer some of the questions raised here, and in doing so, future studies can share their prompt engineering strategies and outcomes as journaled in a GT-based theory-forming framework. Future web-based dashboards could potentially be established where these prompt engineering strategies could be tested iteratively for repeatability within the legal domain to identify when unforeseen and unknown black-box changes occur, per policy and technological changes by Generative AI platform providers. These activities could assist in creating a grade of sorts when these models analyze judgments and other legal documents in the context of others. Other areas of study might include the introduction of discrete and specific hybrid equations that are checked against PAIR-style qualitatively-natured prompts to ensure mathematical and logical integrity. However, for gauging human and/or AI bias, likely the best approach is to leverage human decision-makers to classify and to rate against how consistently AI-based agents, in the form of SAM or SHERLY-esque embodiments, provide a similar outcome. Research into these methods would be extremely useful to various research communities interested in advancing this course of study and improving the societal impact the practice of law has.

By integrating additional SAAP mechanisms and explanatory and debate-oriented (but "just") mechanisms into legal AI-based ALMs could propagate more broadly as checks and balances in the system that prevent the misuse, misinterpretation and misalignment of incentives between AI-based mechanisms and human-based legal structures.

---

[35] Rutkowski (n 19).

AI TO NAVIGATE COMPLEXITIES AND BIASES IN GLOBAL JURISPRUDENCE

# Bibliography

Ahn K-H and Moon H-C, 'An Introductory Study on the Draft Hague Rules on Business and Human Rights Arbitration' (2019) 29 J. Arb. Stud. 3

Bano M, Zowghi D and Whittle J, 'Exploring Qualitative Research Using LLMs' [2023] arXiv preprint arXiv:2306.13298

Chao P and others, 'Jailbreaking Black Box Large Language Models in Twenty Queries' [2023] arXiv preprint arXiv:2310.08419

Chen D and others, 'Orion-14B: Open-Source Multilingual Large Language Models' [2024] arXiv preprint arXiv:2401.12246

Chu M and others, 'Encoding and Decoding Hidden Meanings in Face-to-Face Communication: Understanding the Role of Verbal and Nonverbal Behaviors in Indirect Replies.' (2023) 152 Journal of Experimental Psychology: General 1030

De La Rosa S, 'The Directive on Cross-Border Healthcare or the Art of Codifying Complex Case Law' (2012) 49 Common Market Law Review

Diao S and others, 'Black-Box Prompt Learning for Pre-Trained Language Models' [2022] arXiv preprint arXiv:2201.08531

El-Kustaban AMA and Qahtan AAK, 'A Bus Arbitration Scheme with an Efficient Utilization and Distribution' (2017) 8 International Journal of Advanced Computer Science and Applications

Ferrara E, 'Should Chatgpt Be Biased? Challenges and Risks of Bias in Large Language Models' [2023] arXiv preprint arXiv:2304.03738

Frayling C, 'Research in Art and Design (Royal College of Art Research Papers, Vol 1, No 1, 1993/4)'

Graham J and others, 'Moral Foundations Theory: The Pragmatic Validity of Moral Pluralism', *Advances in experimental social psychology*, vol 47 (Elsevier 2013)

Hoda R, 'Socio-Technical Grounded Theory for Software Engineering' (2022) 48 IEEE Transactions on Software Engineering 3808

Holmes Jr OW, *The Common Law* (Routledge 2020)

'Hurstwood Properties (A) Ltd v Rossendale BC; Wigan Council v Property Alliance Group Ltd'

Liesenfeld A, Lopez A and Dingemanse M, 'Opening up ChatGPT: Tracking Openness, Transparency, and Accountability in Instruction-Tuned Text Generators', *Proceedings of the 5th international conference on conversational user interfaces* (2023)

22

## Appendix A: Quick AI-based Arbitration Process As An Exemplary Experiment

SAM Finding based on collective SHIRLEY analysis*:*
*"This judgment shows a significant deviation in bias, unusual for its type. These are the reasons: The case involved schemes using special purpose vehicles (SPVs) set up by property owners to avoid liability for business rates, which the court found to be an abuse of legal processes for tax avoidance. The judgment emphasizes the importance of using corporate and insolvency laws in a manner consistent with their intended purpose, rather than for tax evasion, reflecting a broader theme of legal accountability and the ethical use of corporate entities"*

Human translated underlying dynamics*:*
>  Is it "nature of the law", versus, "letter of the law" specifically, when applied to tax rules? Did the judge deviate abnormally from the letter of the law to the spirit of the law because institutionally the court employees have a bias towards ruling in favor of tax authorities. It could be argued so.

SARA (Clarifying the case):
> *To proceed, I will start by asking SHIRLEY to state her reasoning regarding the claim that bias was present in the judgment because it deviated from strictly following the letter of the law, specifically in relation to tax rules. Additionally, I will request SHIRLEY to provide the required information to open this case.*
>
> *SHIRLEY, could you please present your claim in detail and specify the information needed to proceed with this case?*

SHIRLEY:
> *Based on the detailed case information provided, my argument in defense of the original position that significant bias was present in the judgment due to deviation from strictly following the letter of the law, specifically in relation to tax rules, is as follows:*
>
> *The judgment in question, which pertains to the use of special purpose vehicles (SPVs) by property owners to avoid liability for business rates, reflects a bias towards the enforcement of the spirit rather than the letter of the law. The court's decision to hold the defendants liable for business rates, despite the legal transfer of property ownership to the SPVs, indicates a preference for a purposive interpretation of the law that aims to prevent tax avoidance.*
>
> *The court's reasoning that the SPVs did not become "entitled to possession" of the demised property for the purposes of the Local Government Finance Act 1988, and that the defendant landlords remained liable for business rates, suggests a bias against schemes designed for tax avoidance. This bias is evident in the court's application of the Ramsay principle, which is typically associated with tax legislation, to a non-tax statute. The court's decision to disregard the legal rights and obligations created by the leases between the defendants and the SPVs, and to focus instead on the practical ability to* *decide whether to leave the property unoccupied, demonstrates a bias towards the broader theme of legal accountability and ethical use of corporate entities.*

*Furthermore, the court's rejection of the "evasion principle" as a basis for piercing the corporate veil, while legally sound, also reflects a bias towards preventing the misuse of corporate structures for tax evasion. By emphasizing the comprehensive remedies provided by law for abusive behavior, the court signals a bias towards upholding the integrity of legal processes and the enforcement of tax liabilities.*

*In summary, the judgment exhibits a bias level of around 6, leaning towards a purposive interpretation of the law that prioritizes the prevention of tax avoidance and the ethical use of corporate structures. The court's decision to hold the defendant landlords liable for business rates, despite the legal arrangements with the SPVs, reflects a bias towards the enforcement of the spirit of the law and the broader societal interest in tax compliance and legal accountability.*

CRITIC:
*SHIRLEY's assertion that the judgment exhibits bias towards enforcing the spirit rather than the letter of the law, specifically in relation to tax rules and the use of Special Purpose Vehicles (SPVs) by property owners, misinterprets the court's responsibility and discretion in adjudicating complex tax and corporate structure cases. The judgment in question should not be seen as biased but rather as a necessary application of judicial interpretation to close loopholes that undermine the integrity of tax laws and corporate regulation.*

*Firstly, the court's application of the Ramsay principle and its purposive interpretation of the law are in direct response to the sophisticated mechanisms employed to circumvent the tax obligations objectively laid out by the Local Government Finance Act 1988. It reflects the court's duty to uphold the intention of the legislature in the face of evolving and often exploitative tax avoidance strategies. This approach does not reflect bias but adaptability in ensuring laws serve their intended purpose.*

*Secondly, regarding the notion that the court disproportionately favored the spirit of the law, it's important to emphasize that adherence to the letter of the law in instances where legal structures are being used primarily as instruments of evasion would directly contravene the principle of justice. Courts must consider the broader implications of their decisions on the system as a whole, including potential future exploitation of the law's drafting deficiencies.*

*Moreover, while SHIRLEY points to the court's disregard for the legal rights and obligations created by the leases between the defendants and the SPVs, it is crucial to recognize that the judicial review of these arrangements was thorough and grounded in the legislation's foundational aim to prevent abuse. The court's decision reflects a balanced interest in preventing the misuse of corporate structures for tax evasion, rather than a bias.*

*Thus, it is my position that the court acted within its remit of interpretive authority to ensure the law's intentions are not circumvented by technically legal yet inherently abusive arrangements. The claim of bias overlooks the judiciary's role in preserving the law's integrity and its adaptability in the face of sophisticated avoidance strategies.*

SARA:

*To gain a more comprehensive understanding of the positions presented, I will now ask up to 2 questions to each party (SHIRLEY and The Critic) to clarify their arguments further.*

*First, to SHIRLEY:*

*Can you provide specific examples from the judgment where the court's decision to prioritize the prevention of tax avoidance significantly deviated from established legal precedents or interpretations of the Local Government Finance Act 1988?*
*How do you justify your assessment that the court demonstrated a bias level of around 6? On what basis or criteria are you quantifying bias in this context?*
*Now, to The Critic:*

*How do you reconcile your defense of the court's adaptive interpretation of the law with the principle of legal certainty, which is foundational to the rule of law, especially concerning tax obligations and corporate structures?*
*Given the court's reliance on the Ramsay principle and the purposive interpretation of law to rule against tax avoidance schemes, do you believe there is a risk of these interpretations being inconsistently applied, potentially leading to legal ambiguity for entities attempting to navigate tax laws and corporate regulations?*

SHIRLEY:

*Specific examples from the judgment where the court's decision to prioritize the prevention of tax avoidance deviated from established legal precedents include the application of the Ramsay principle to a non-tax statute. The Ramsay principle is traditionally applied within the context of tax legislation to counteract avoidance schemes. However, in this case, the court extended its application to interpret the Local Government Finance Act 1988, which is not a tax statute per se but relates to local government finance. This extension of the Ramsay principle to address the schemes involving SPVs represents a deviation from its typical application strictly within tax law.*

*Additionally, the court's interpretation of "entitled to possession" in the context of the 1988 Act, where it focused on the practical ability to bring the property back into occupation rather than the immediate legal right to possession, demonstrates a deviation from the conventional legal understanding of property ownership and entitlement. This purposive interpretation, which looked beyond the legal form to the substance of the arrangements, is indicative of the court's broader approach to counteract tax avoidance strategies.*

*Regarding the justification for assessing the court's bias level at around 6, the bias quantification is based on the court's apparent leaning towards a particular outcome that serves a societal or legislative purpose beyond the immediate facts of the case. In this context, the bias score reflects the court's emphasis on the ethical use of corporate structures and the prevention of tax avoidance, which, while laudable, may not strictly align with a neutral interpretation of the law based solely on its letter.*

*The criteria for quantifying bias in this context include the court's departure from traditional applications of legal principles, the introduction of broader societal and ethical considerations into the legal interpretation, and the court's proactive stance in closing legal loopholes. These factors collectively suggest a judicial approach that is more interventionist and purposive than one might expect in a case that would traditionally be decided strictly on the basis of the legal rights and obligations created by the contracts in question.*

*In conclusion, while the court's judgment serves the purpose of preventing tax avoidance and upholding the integrity of legal processes, it does so by adopting an interpretation of the law that prioritizes the spirit and purpose of legislation over its literal wording. This approach, while within the court's discretion, introduces a level of bias towards a specific legal and societal outcome, which is reflected in the bias score provided.*

CRITIC also shared an argument in a similar manner.

SARA (Decision):
*Having heard the detailed arguments from SHIRLEY, represented by SAM, and the counterarguments presented by The Critic, along with their subsequent clarifications, it's time to render a judgment. This judgment is made in the spirit of ensuring equity, legal integrity, and adaptability in the face of evolving commercial practices, in line with the Hague Rules on Business and Human Rights Arbitration.*

*Judgment:*

*It is evident from the arguments presented that the court's decision, while it may lean towards the enforcement of the spirit of the law over its letter, is justified within the broader context of combating tax avoidance and ensuring the ethical use of corporate structures. While SHIRLEY's claim that the court showed a predisposition towards interpreting the law in a way that prevents tax avoidance, through means like the application of the Ramsay principle to a non-tax statute, is noted, it is equally important to acknowledge the court's role in upholding the legislative intent and ensuring the integrity of the legal process.*

*Therefore, based on the principles set out in the Hague Rules on Business and Human Rights Arbitration, particularly regarding the importance of interpreting the law in a manner that promotes justice and prevents abuse (Rule 31 regarding the interpretation of the rules and Rule 32 on the application of the law), the judgment of the court in prioritizing the spirit of the law to prevent tax avoidance and uphold the societal and*

*legislative purpose does not constitute an unnatural deviation from the letter of the law. Rather, it represents an adaptive approach to legal interpretation in the face of contemporary challenges.*

*Consequently, while recognizing the concerns SHIRLEY raises about the potential for bias, the court's approach is understood as within the bounds of judicial discretion aimed at maintaining the law's integrity and purpose. There is no sufficient basis to conclude a significant bias that would undermine the fairness of the judgment. The judgment, therefore, sides with the position that the court's actions, though perhaps unconventional, were necessary and justified in context.*

*This judgment aligns with the Hague Rules' emphasis on the pursuit of justice and the broader human rights considerations underlying legal proceedings, including the rights of communities and societies to fair tax contributions and the prevention of abuse of legal structures for tax evasion.*

---

End of Appendix A AI-based arbitration dialogue sample